%% file: main.tex
\documentclass[conference]{IEEEtran}
\IEEEoverridecommandlockouts
\usepackage{cite,footmisc}
\usepackage{amsmath,amsthm,amssymb,amsfonts,bm,mwe}
\usepackage{algorithmic}
\usepackage{graphicx,subcaption,subfloat}
\usepackage{textcomp}
\usepackage{xcolor,url}
\usepackage{tabularx,caption}
\usepackage{rotating,multirow,booktabs}
\usepackage{nicefrac,upgreek}
\usepackage{flushend}
\def\BibTeX{{\rm B\kern-.05em{\sc i\kern-.025em b}\kern-.08em
    T\kern-.1667em\lower.7ex\hbox{E}\kern-.125emX}}
\def\code#1{\texttt{#1}}
\newcommand{\RN}[1]{%
  \textup{\uppercase\expandafter{\romannumeral#1}}%
}
\begin{document}

\title{Gaussian Function On Response Surface Estimation\\
}

\author{\IEEEauthorblockN{1\textsuperscript{st} Mohammadhossein Toutiaee}
\IEEEauthorblockA{\textit{Computer Science Department} \\
\textit{The University of Georgia}\\
Athens, Georgia, U.S. \\
hossein@uga.edu}
\and
\IEEEauthorblockN{2\textsuperscript{nd} John A. Miller}
\IEEEauthorblockA{\textit{Computer Science Department} \\
\textit{The University of Georgia}\\
Athens, Georgia, U.S. \\
jamill@uga.edu}
}


\maketitle

\begin{abstract}
\input{tex/0-abstract}

\end{abstract}

\begin{IEEEkeywords}
Machine Learning Interpretation, Metamodeling, Surrogate Models, Explainable Artificial Intelligence, Gaussian Process
\end{IEEEkeywords}

\section{Introduction}
\input{tex/1-Intro}


\section{Gaussian Function On Response Surface Estimation; G-FORSE}
\input{tex/3-gforse}


\section{Experiments}
\input{tex/5-exp}

\section{Advantages and Disadvantages}
\input{tex/6-advantage}

\section{Conclusions and Future Work}
\input{tex/7-conc}

\bibliographystyle{IEEEtran}
\bibliography{IEEEabrv,main}

\end{document}

%% file: tex/0-abstract.tex
We propose a new framework for 2-D interpreting (features and samples) black-box machine learning models via a metamodeling technique, by which we study the output and input relationships of the underlying machine learning model.
The metamodel can be estimated from data generated via a trained complex model by running the computer experiment on samples of data in the region of interest.
We utilize a Gaussian process as a surrogate to capture the response surface of a complex model, in which we incorporate two parts in the process: interpolated values that are modeled by a stationary Gaussian process $Z$ governed by a prior covariance function, and a mean function $\mu$ that captures the known trends in the underlying model.
The optimization procedure for the variable importance parameter $\theta$ is to maximize the likelihood function.
This $\theta$ corresponds to the correlation of individual variables with the target response.
There is no need for any pre-assumed models since it depends on empirical observations.
Experiments demonstrate the potential of the interpretable model through quantitative assessment of the predicted samples.

%% file: tex/1-Intro.tex
With the prompt acceptance of sophisticated Artificial Intelligence (AI) models in the industry for solving problems, Machine Learning Interpretation (MLI) is not just a new research direction but a need.
Model explanation enables us to interpret and legitimize the outcomes of a predictive model to accredit Fairness, Accountability and Transparency (FAT) in making decisions.
In other words, the theory and mechanics of algorithmic decisions should be deciphered such that the relationship between inputs and outputs are understandable to humans. 

Determining FAT in predictive models is a major problem when a model is used for decision making. 
When using a complex gradient boosting model (GBM) with rich parameters to predict membership, for example, prediction cannot be acted upon an interpretable algorithmic process due to the fact that thousands of pre-trained regression trees seem like a black-box to humans. 

Apart from understanding the black-box process, the complex model should have variable relationships evaluated before launching it into the production.
To make this decision, humans require to validate what has actually been trained via an algorithmic model to what they perceive from that model.
That is, they will not expect to see correlations between a group of variables and a target in a prediction problem when the model has excluded that group from the prediction process.
In general, machine learning interpretation is evaluated based on Local Inference (single instance) and Global Inference (entire or part of data).
However, real-world models are often significantly complex, and further, we do not have enough understanding of why and when they work well, and why they may fail completely when faced with new situations not seen in the training data.
A transparent and simplified version (a surrogate or metamodel) of a complex model is an effective solution, in addition to such methods.
In this case, a metamodel provides the best approximation to the underlying model by minimizing a metamodeling loss $\ell (g,f)$, where $g$ and $f$ are the metamodel and black-box, respectively.

\subsection{Metamodeling}
\textit{Metamodeling} is a branch of computer experiments in design of experiments (DOE), where a metamodel can be estimated by sampling data points from a complex model that has been trained expensively by using extensive resources. 
Metamodeling (or surrogate) is utilized when an outcome of interest in a complex model cannot be directly identified easily.
In this situation, traditional DOE methods would fail to capture the structure and relationships (geometry) of the model.
Modeling geometric effects is vital when the purpose of the experiment is a study of how combinations of various inputs influence an output (response) measuring quantity or quality characteristics.
This enables us to identify the relationship between the input (predictors $X$) and output (response $y$) variables in a complex machine learning process, by fitting a global metamodel to the underlying model.
The aim of \cite{kianifar2020performance} was to present a fair evaluation on several metamodels widely used by practitioners, comparing different characteristics of the techniques regarding robustness, accuracy, efficiency, etc. 
Given the fact that each metamodel is able to focus on one or two aspects of a model, they showed that not all techniques are powerful in every aspect, so the metamodeling strategy should be selected carefully.
The paper provides a comprehensive evaluation of techniques for each aspect, and it concluded that Gaussian Process could outperform other metamodels in some aspects.

One aspect of metamodeling which has been less studied is ``black-box interpretation'', and this paper aims to provide thorough insights by introducing a new technique for machine learning interpretation through the Gaussian Process method.
A metamodel is useful for interpretation because it is a simplified copy of black-box model. 
A complex prediction model represents a raw data and conforms to a metamodel.

In this article, we propose a metamodel for global interpretation as a solution to enable prediction of FAT, and optimize the metamodel by interpolation to explore the relationships between several explanatory variables and one or more response variables.
Our main contributions are summarized as follows.

\begin{itemize}
    \item The framework that can explore the relationships between several input variables and one or more target variables by metamodeling.
    \item The metamodel that can reveal the structure and relationships between data points, while simplifying the explanations of the complex model.
    \item A surrogate for a simple model such as Logistic Regression model that can suffer from instability of estimation as a result of Hauck-Donner effect.
    \item Comprehensive evaluation of metamodeling in interpreting a complex model on some datasets, where we evaluate the strength of a feature for prediction.
    In our experiments, the final fitted model obtained from our proposed approach is more interpretable and decipherable to the viewers.
    We also show how the mechanics and theory of metamodel characterized by the ``meta'' of metamodeling can be used to emulate the distribution of a black-box efficiently by employing the black-box's resources.
\end{itemize}

%% file: tex/3-gforse.tex
Our work estimates the black-box information through a Kriging process \cite{kaymaz2005application} by defining a model consisting of two parts: linear regression part and non-parametric stochastic part, which can be given as:
\begin{eqnarray}
Y(x) = \mu(x) + Z(x),
\end{eqnarray}
where $\mu(x) = \sum_{i=1}^m q_i(x)\beta_i=q^T(x)\beta$.
$\beta=[\beta_1,\ldots,\beta_m]^T$ is the regression coefficient to be determined, and $q(x)=[q_1(x),\ldots,q_m(x)]^T$ is function of vector $x$ which can provide the global approximation.
$Z(x)$ is a stationary Gaussian stochastic process with mean $0$ and covariance function:
\begin{eqnarray}
\mathbb{C}(x_i,x_j) = Cov(Z(x_i),Z(x_j))=\sigma^2\prod_{l=1}^k K(h_l;\theta_l),
\end{eqnarray}
where $\sigma^2$ is the variance parameter, $h_l = |x_i^{(l)}-x_j^{(l)}|$, $x_i^{(l)}$ and $x_j^{(l)}$ are the $l$th elements of the $i$th run $x_i$ and the $j$th run $x_j$, $k$ is the number of variables and $K(h_l;\theta_l)$ is a correlation function with a positive parameter $\theta_l$.
G-forse can be calibrated by choosing an impactful correlation function, and several alternatives such as cubic, exponential and Mat\'ern functions have been studied in \cite{koehler1996computer}.
Among which, the Gaussian model is utilized in this work \cite{forrester2008engineering}, which it is provided in equation 3.
We aim to approximate the true underlying model $Y(x)$ by the best linear unbiased estimator (BLUE), which minimizes an objective function $\mathbb{E}\{\hat{Y}(x)-Y(x)\}^2$, under the model in equation 1.
The function $\mu$ is used to identify the known trends in the equation, so it enables $Z(x)$ to be a stationary process.
\textit{Ordinary Kriging} takes $\mu$ in equation 1 as constant $\mu_0$, which is widely used in studies \cite{wang2013application, forrester2008engineering,welch1992screening}\nocite{toutiaee2020video}.

The G-forse framework is most straightforward to apply when the basis function is Gaussian product correlation of the form
\begin{equation}
\psi(\bm{h}) = exp\left(-\sum^k_{j=1}\theta_j h^{p_j}\right)    
\end{equation}
This function is very similar to a Gaussian process where the observations have Gaussian basis function described with two characteristics functions $\mu(x)$ (mean) and $\mathbb{C}(x,x')$ (covariance).
We intend to use a vector $\bm{\theta} = \{\theta_1,\theta_2,\ldots,\theta_k\}^T$ in the G-forse to control the width of the basis function varying from feature to feature, while the Gaussian basis function has $1/\sigma^2$.
Similarly, the Gaussian kernel tends to use a fixed exponent at $p=2$ to enable smooth function in all dimensions for point $x^{(i)}$.
The Gaussian process is a special case of the metamodeling process when $\theta_j$ is fixed and $p_{(1,2,\ldots,k)}=2$ for all dimensions (isotropic basis function).
In the next section, we present a theoretical analysis of G-forse process, essentially showing that how smoothness and activeness of $\bm{p}$ and $\bm{\theta}$, respectively, affect the underlying correlation.
In practice, we must view our observed responses $\bm{y}=\{y^{(1)},y^{(2)},\ldots,y^{(n)}\}^T$ as if they are sampled from a stochastic process, although they may be deterministic in code.
Thus, our observed responses are denoted by $Y=\{Y(x^{(1)}),Y(x^{(2)}),\ldots,Y(x^{(n)})\}$ with a mean of {\boldmath $1$}$\mu$.

Practically, we assume that $cor[Y(\bm{x}^{(i)}),Y(\bm{x}^{(l)})]$ reflects our expectation function (equation 3) and it is smooth and continuous in the defined space.
Such assumptions provide some correlations between a set of random variables {\boldmath $Y$} that are relying on parameters $\theta _j$ and $p$ and the distance between points $|x_j^{(i)}-x_j^{(l)}|$.
The likelihood function, which can be expressed in terms of the sample data, provides us the concentrated log-likelihood function which optimize the locations of unknown parameters, and consequently, it enables us to determine the rank of importance of variables.

%% file: tex/5-exp.tex
We demonstrate the use cases of G-forse through experiments on synthetic and real data. 
In all experiments, we used \code{SPOT} Gaussian Process computation library in R to carry out computations involving G-forse.

\begin{figure*}[htb!]
        \centering
        \begin{minipage}{0.48\linewidth}
        \begin{subfigure}[b]{0.48\linewidth}
            \centering
            \includegraphics[width=\linewidth]{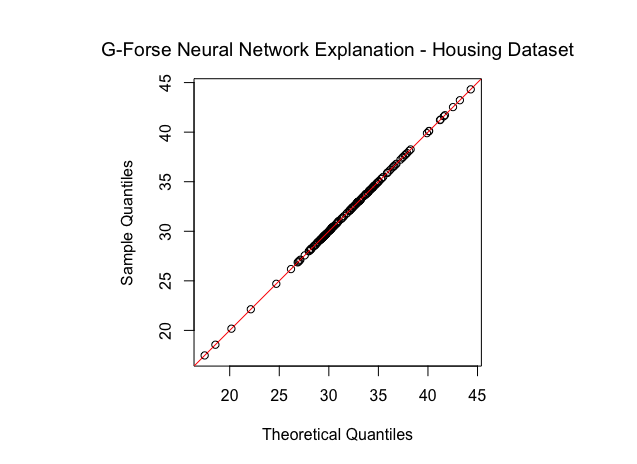}
            \caption{}
            \label{fig:1a}
        \end{subfigure}
        \hfill
        \begin{subfigure}[b]{0.48\linewidth}
            \centering
            \includegraphics[width=\linewidth]{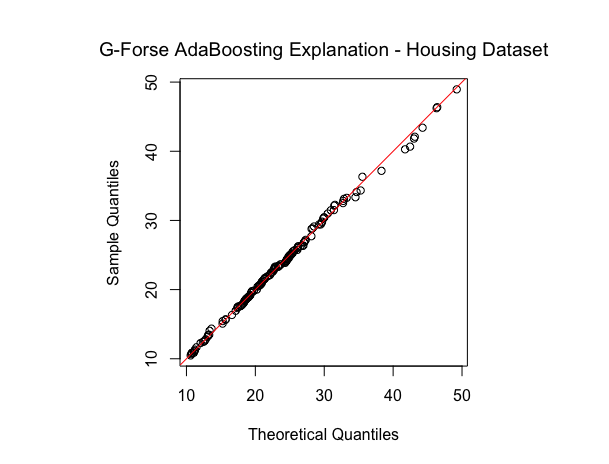}
            \caption{}
            \label{fig:1b}
        \end{subfigure}
        \vskip\baselineskip
        \begin{subfigure}[b]{0.48\linewidth}
            \centering
            \includegraphics[width=\linewidth]{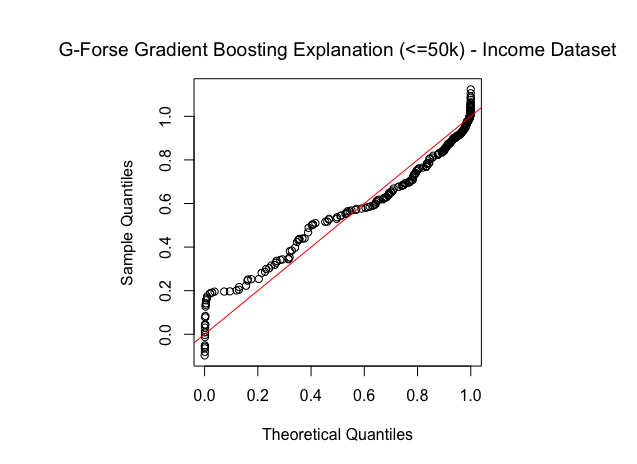}
            \caption{}
            \label{fig:1c}
        \end{subfigure}
        \hfill
        \begin{subfigure}[b]{0.48\linewidth}
            \centering
            \includegraphics[width=\linewidth]{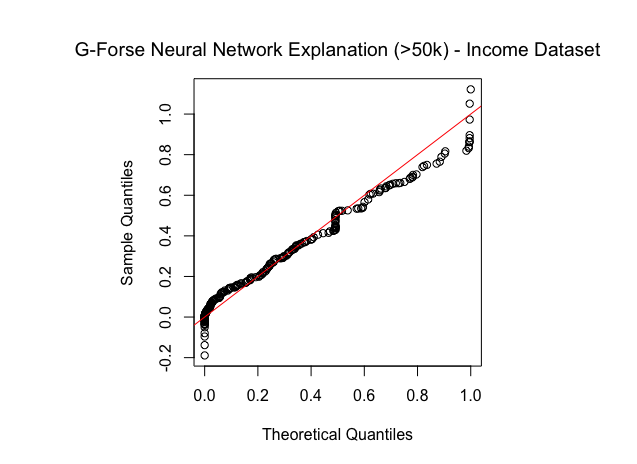}
            \caption{}
            \label{fig:1d}
        \end{subfigure}
        \caption{Scatter plots of regression outcomes computed by G-forse versus outcome values produced by different black-box models across various datasets.}
        \label{fig:qqreg}
  \end{minipage}
  \hfill
  \begin{minipage}{0.48\textwidth}
        \begin{subfigure}[b]{0.48\linewidth}
            \centering
            \includegraphics[width=\linewidth]{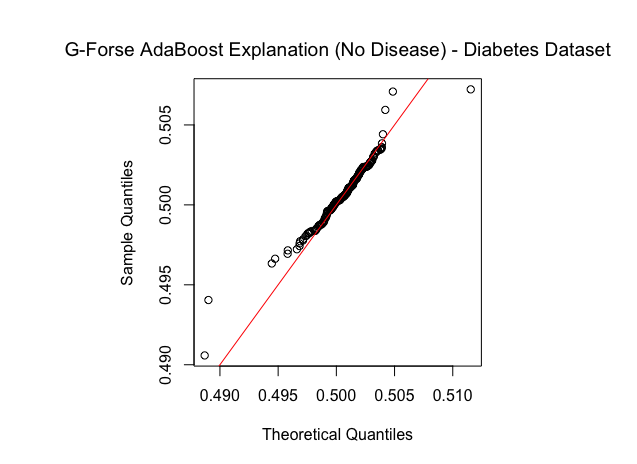}
            \caption{}
            \label{fig:1e}
        \end{subfigure}
        \hfill
        \begin{subfigure}[b]{0.48\linewidth}
            \centering
            \includegraphics[width=\linewidth]{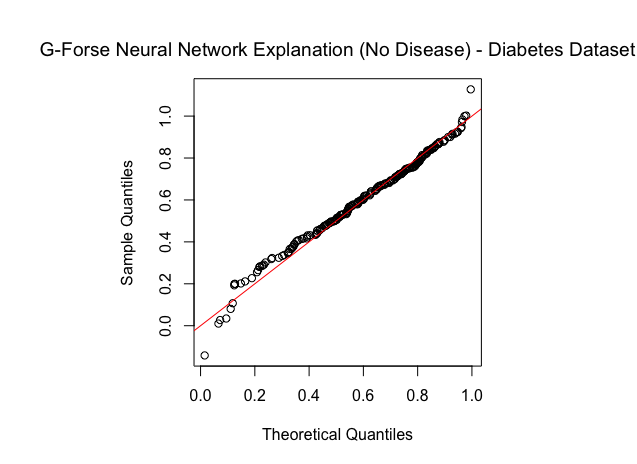}
            \caption{}
            \label{fig:1f}
        \end{subfigure}
        \vskip\baselineskip
        \begin{subfigure}[b]{0.48\textwidth}
            \centering
            \includegraphics[width=\linewidth]{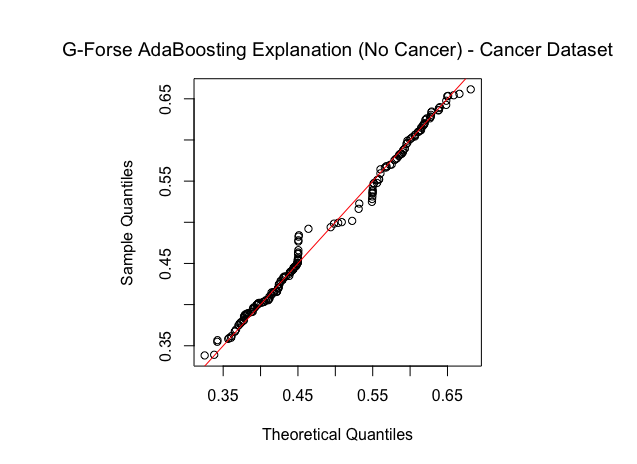}
            \caption{}
            \label{fig:1g}
        \end{subfigure}
        \hfill
        \begin{subfigure}[b]{0.48\textwidth}
            \centering
            \includegraphics[width=\linewidth]{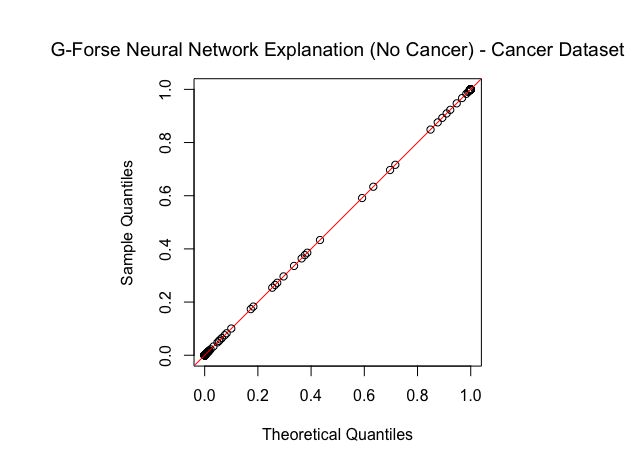}
            \caption{}
            \label{fig:1h}
        \end{subfigure}
        \caption{Scatter plots of classification outcomes computed by G-forse versus outcome values produced by different black-box models across various datasets.}
        \label{fig:qqclass}
  \end{minipage}
    \end{figure*}

We pretend the validated prediction of a trained complex model on a dataset is a new target variable, which G-forse aims to predict. 
This enables G-forse to provide the global explanation of the complex model after G-forse has been enforced to predict the new target (validated prediction) using original inputs.
We trained four popular machine learning models on a range of datasets, and we tested G-forse on the validated outputs of those algorithms (Tables \ref{tab:de} \& \ref{tab:bfgs}).
We also tested G-forse on a simulated data generated from a true GLM model with pre-defined weights and correlations to ensure it performs trustworthily.
Among which, the crimes and housing price datasets are regression problems with the continuous response variables, and the remaining ones are the classification problems with the binary response variables.
For fitting a G-forse model, the values for each variable are normalized to 0 to 1 range.
We only tuned parameter $\hat{\theta}$ since it determines the variable importance of a model, which we desire.
While our theoretical framework permits the tuning of parameter $\hat{p}$ and one can benefit from optimizing $\hat{p}$ to produce accurate predictions, the parameter $\hat{p}$ was fixed at $p=2$ value, because we have a smooth
correlation with a continuous gradient for very close points.
We estimate parameter $\hat{\theta}$ by use of differential evolution (DE) and L-BFGS-B optimizers \cite{storn1995differrential,byrd1995limited}, and we limit the search region by setting the lower and upper bounds to $10^{-4}$ and $10^2$, respectively.
The results are provided in Tables \ref{tab:de} \& \ref{tab:bfgs} for comparison.
Superficially, G-forse with DE optimizer achieved similar results to L-BFGS-B optimizer: Table \ref{tab:bfgs} differs from Table \ref{tab:de} in that L-BFGS-B optimizer worked better on the Income dataset.
However, both tables are showing G-forse performed reasonably well in approximating the most algorithms across all the datasets. 
The results from both tables are in line with the results on probability plots we presented in Figures \ref{fig:qqreg} and \ref{fig:qqclass}, where they are showing that G-forse is successfully able to approximate the underlying predictive models.

\subsection{Results}
The prediction of G-forse model is obtained by Gaussian process with zero mean and covariance matrix by affine transformation of correlation matrix $\Psi$ using Cholesky decomposition technique. The performance is measured via the root mean squared error ($RMSE$) and the correlation ($r$) between predicted and produced observations from the underlying model (Table \ref{tab:de} and \ref{tab:bfgs}).
In the G-forse prediction process, the errors $\epsilon(x^{(i)})$ are the realization of Gaussian process, and we predict $\hat{y}$ at unknown location $x^*$ by including $\hat{y}$ into the known observations $\Tilde{y}=\{y,\hat{y}\}^T$. We treat this as the model parameter which is estimated by use of MLE.
A G-forse process model is parsimonious and appropriate for high dimensional data, while keeping the number of estimated parameters very low.
While we make no claim that G-forse model can outperform other existing interpretable methods in explaining and performance, we believe that G-forse outperforms the GLMs such as Logistic Regression in interpretation, overcoming the issue of Hauck-Donner effect, and it can be competitive with the methods in the literature and highlight the potential of the G-forse model.

\subsection{Group Explanation}
\begin{figure*}
       \centering
\begin{subfigure}[b]{0.24\textwidth}
\includegraphics[width=\textwidth]{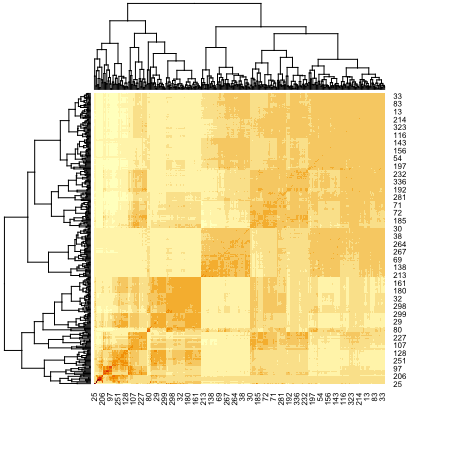}
\caption{Samples in AdaBoost \\ \centering(Housing data)}
\end{subfigure}
\begin{subfigure}[b]{0.24\textwidth}
\includegraphics[width=\textwidth]{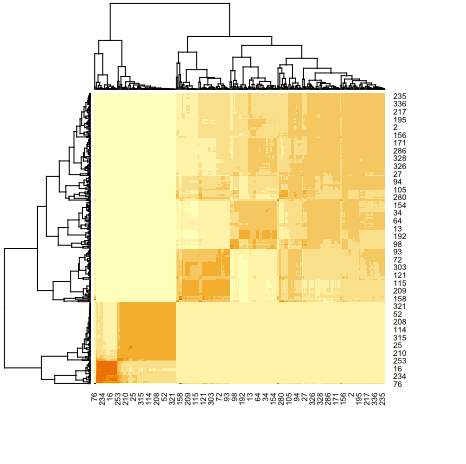}
\caption{Samples in Neural Network \\ \centering(Housing data)}
\end{subfigure}
\begin{subfigure}[b]{0.24\textwidth}
\includegraphics[width=\textwidth]{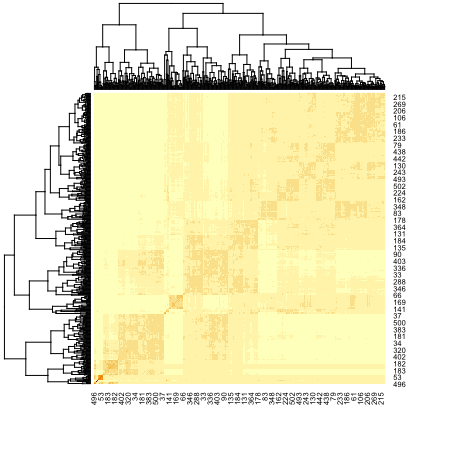}
\caption{Samples in Neural Network \\ \centering(Diabetes data)}
\end{subfigure}
\begin{subfigure}[b]{0.24\textwidth}
\includegraphics[width=\textwidth]{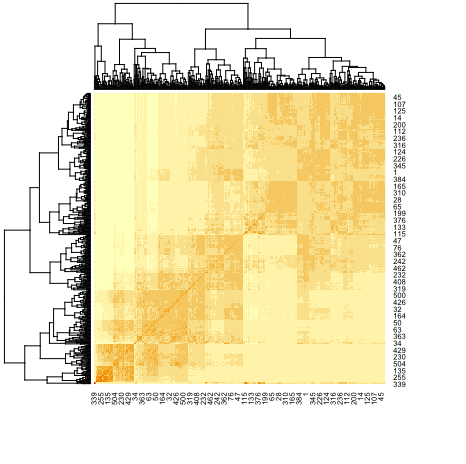}
\caption{Samples in AdaBoost \\ \centering{(Diabetes data)}}
\end{subfigure}

\begin{subfigure}[b]{0.24\textwidth}
\includegraphics[width=\textwidth]{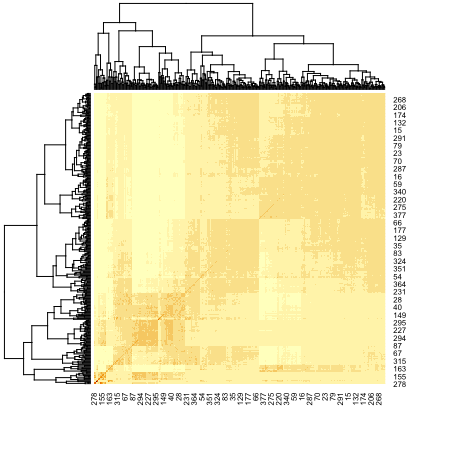}
\caption{Samples in AdaBoost \\ \centering(Cancer data)}
\end{subfigure}
\begin{subfigure}[b]{0.24\textwidth}
\includegraphics[width=\textwidth]{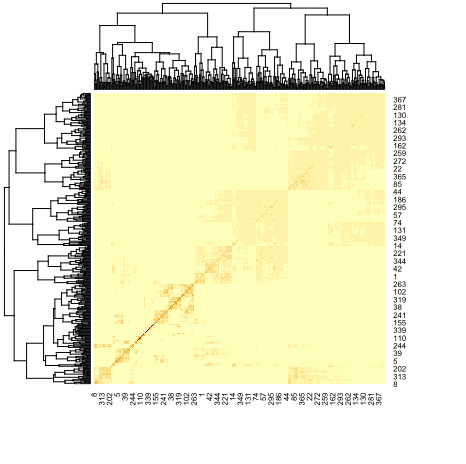}
\caption{Samples in Neural Network \\ \centering(Cancer data)}
\end{subfigure}
\begin{subfigure}[b]{0.24\textwidth}
\includegraphics[width=\textwidth]{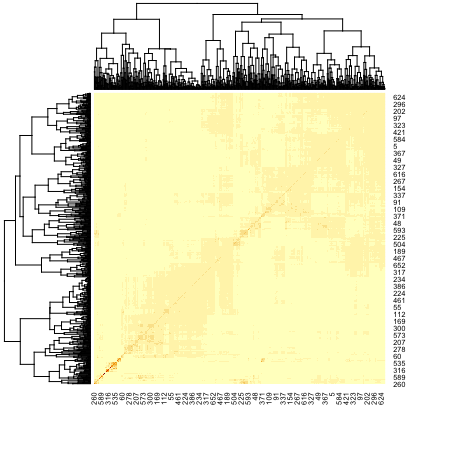}
\caption{Samples in Gradient Boosting \\ \centering(Income data)}
\end{subfigure}
\begin{subfigure}[b]{0.24\textwidth}
\includegraphics[width=\textwidth]{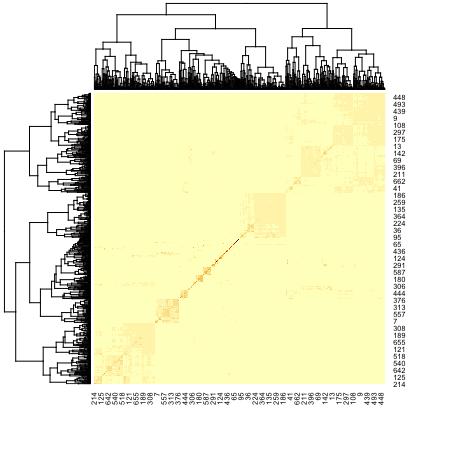}
\caption{Samples in Neural Network \\ \centering{(Income data)}}
\end{subfigure}
\caption{Clustering between data points revealed by G-forse for different black-box models. These clusters are computed by correlation matrix estimated during G-forse optimization. Each machine learning model treats samples differently in predicting the outcome.}\label{fig:heat}
    \end{figure*}

Of interest to practitioners is to find if global interpretation is appropriate for demystifying the target black-box.
Conventionally, MLI techniques deliver variable selection by computing feature importance based on some metric function, so they determine what group of variables are important for prediction.
What is less studied in the field of MLI is knowing what \textit{group of samples} are contributing together to the prediction made by the black-box, and G-forse undertakes this idea by constructing a \textit{correlation} function during the training phase (Figure \ref{fig:heat}).
The correlation function ($\bm\Psi$) used in G-forse provides the benefit to explore correlations between data points given the predictions, which is reflected in matrix $\bm\Psi$.
These correlations are relying on the absolute distance between data points and the parameters $p_i$ and $\theta_j$ in G-forse method.
Figure \ref{fig:heat} shows that if instances $i$ and $j$ are both above or below their respective predictions' means, then the plot shows higher correlation denoted by a darker color.
If one is above its prediction's mean and the other is below, we see lighter (yellowish) spots denoting little to no correlation. 
In other words, the darker areas in the plot indicate an \textit{agreement} on prediction, and the lighter areas mean \textit{disagreement}.
In the classification problems such as (e) or (f) in Figure \ref{fig:heat}, when the outcome is binary (0 or 1), the darker areas represent instances that are in accordance with the prediction made by the black-box (e.g. AdaBoost or Neural Network in (e) and (f)).
This aspect of interpretation made by G-forse provides us with the ability to penetrate inside the black-box and monitor the prediction effect across samples.
Another benefit of group explanation is that the global fidelity of a complex model can be visually appreciated by looking at the heatmap plot (Figure \ref{fig:heat}) to find if it can expand its prediction to more darker areas.
In this case, the user might choose between a global or local explanation depending on how the prediction was made by the underlying model.
\begin{table*}[t]
    \centering
        \begin{tabular}{c c c c c c c}
            \toprule
            \midrule
                & & \multicolumn{5}{c}{Algorithms - RMSE(r) }\\ \cmidrule{3-7}
                && Neural Network  & GBM & AdaBoost & XGBoost &  \\ \cmidrule{3-7}
                \multicolumn{1}{c}{\multirow{5}{*}{\begin{sideways}\parbox{2cm}{\centering Datasets}\end{sideways}}}   &
                \multicolumn{1}{l}{Crimes}& 0.100(0.88) & 0.101(0.88) & \textbf{0.042(0.95)} & 0.113(0.85)   \\
                \multicolumn{1}{c}{}    &
                \multicolumn{1}{l}{Housing}& \textbf{0.022(0.99)} & \textbf{1.44(0.99)} & \textbf{0.654(0.99)} & \textbf{1.48(0.98)} \\
                \multicolumn{1}{c}{}    &
                \multicolumn{1}{l}{Cancer} & \textbf{0.0001(0.99)} & \textbf{0.126(0.96)} & \textbf{0.015(0.98)} & \textbf{0.072(0.98)}   \\
                \multicolumn{1}{c}{}    &   
                \multicolumn{1}{l}{Diabetes} & \textbf{0.057(0.96)} & 0.361(0.60) & 0.001(0.72) & 0.320(0.66)  \\
                \multicolumn{1}{c}{}    &
                \multicolumn{1}{l}{Income} & \textbf{0.107(0.94)} & 0.154(0.87) & \textbf{0.0002(0.90)} & 0.157(0.86)  \\
            \midrule
            \bottomrule
        \end{tabular}
        \centering
        \caption{The root mean squared error (RMSE) and correlation (r) obtained via G-forse by use of differential evolution optimizer for different complex models on different datasets. Bold values denote significant performance.}
        \label{tab:de}
\end{table*}

\begin{table*}[t]
    \centering
        \begin{tabular}{c c c c c c c}
            \toprule
            \midrule
                & & \multicolumn{5}{c}{Algorithms - RMSE(r) }\\ \cmidrule{3-7}
                && Neural Network  & GBM & AdaBoost & XGBoost &  \\ \cmidrule{3-7}
                \multicolumn{1}{c}{\multirow{5}{*}{\begin{sideways}\parbox{2cm}{\centering Datasets}\end{sideways}}}   &
                \multicolumn{1}{l}{Crimes}& 0.101(0.88) & 0.112(0.86) & \textbf{0.045(0.94)} & 0.110(0.86)  \\
                \multicolumn{1}{c}{}    &
                \multicolumn{1}{l}{Housing}& \textbf{0.259(0.99)} & \textbf{2.26(0.96)} & \textbf{2.13(0.96)} & \textbf{2.40(0.96)} \\
                \multicolumn{1}{c}{}    &
                \multicolumn{1}{l}{Cancer} & \textbf{0.145(0.93)} & \textbf{0.167(0.93)} & \textbf{0.022(0.97)} & \textbf{0.115(0.97)}   \\
                \multicolumn{1}{c}{}    &   
                \multicolumn{1}{l}{Diabetes} & \textbf{0.059(0.96)} & 0.386(0.57) & 0.001(0.74) & 0.336(0.63)  \\
                \multicolumn{1}{c}{}    &
                \multicolumn{1}{l}{Income} & \textbf{0.094(0.94)} & \textbf{0.131(0.91)} & \textbf{0.0002(0.93)} & \textbf{0.144(0.90)}  \\
            \midrule
            \bottomrule
        \end{tabular}
        \centering
        \caption{The root mean squared error (RMSE) and correlation (r) obtained via G-forse by use of L-BFGS-B optimizer for different complex models on different datasets. Bold values denote significant performance. }
        \label{tab:bfgs}
\end{table*}


\begin{figure*}
       \centering
\begin{subfigure}[b]{0.48\textwidth}
\includegraphics[width=\textwidth]{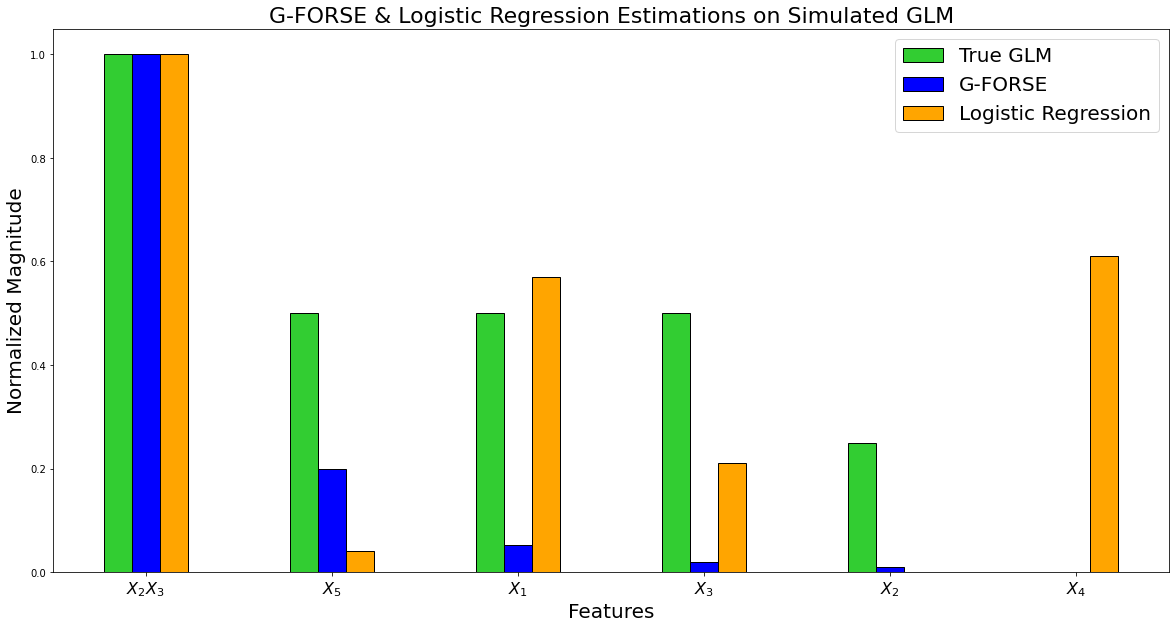}
\caption{}\label{fig:g_lr}
\end{subfigure}
\begin{subfigure}[b]{0.48\textwidth}
\includegraphics[width=\textwidth]{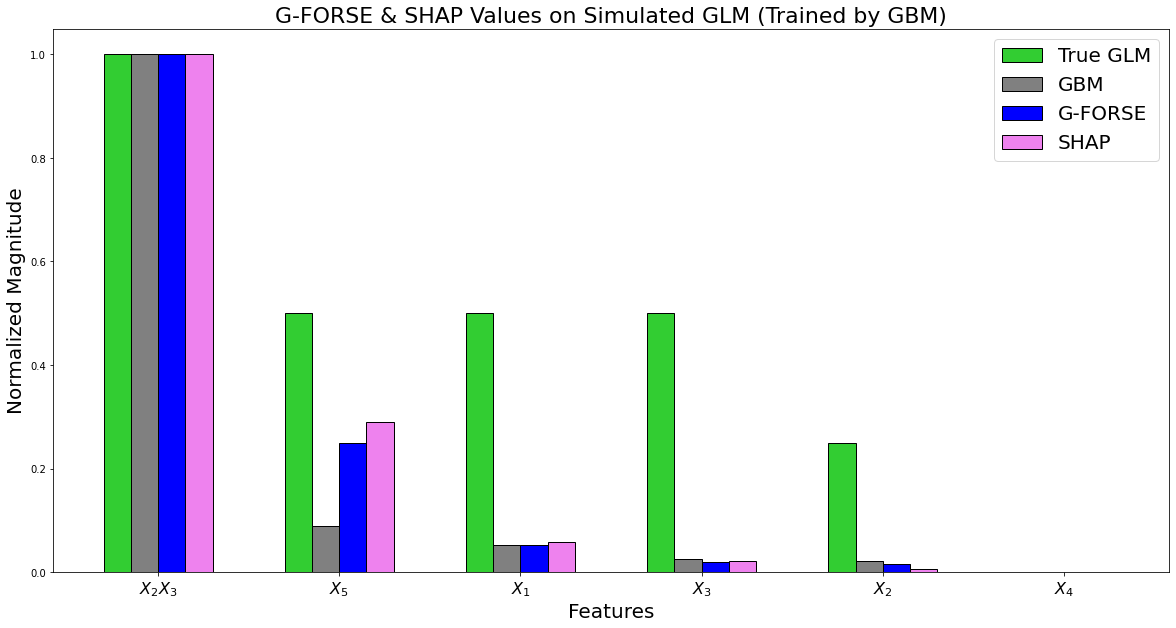}
\caption{}\label{fig:glmShap}
\end{subfigure}
\caption{(a) Side-by-side comparison of features' importance evaluated by G-forse and Logistic Regression on the synthesized GLM. G-forse is superior to Logistic Regression in detecting true signals. (b) G-forse and SHAP values on the synthesized GLM trained by GBM. Both plots show the tendency of G-forse to be trustworthy to approximate global effects.}
\label{fig:barplots}
    \end{figure*}

\begin{figure*}
       \centering
\begin{subfigure}[b]{0.24\textwidth}
\includegraphics[width=\textwidth]{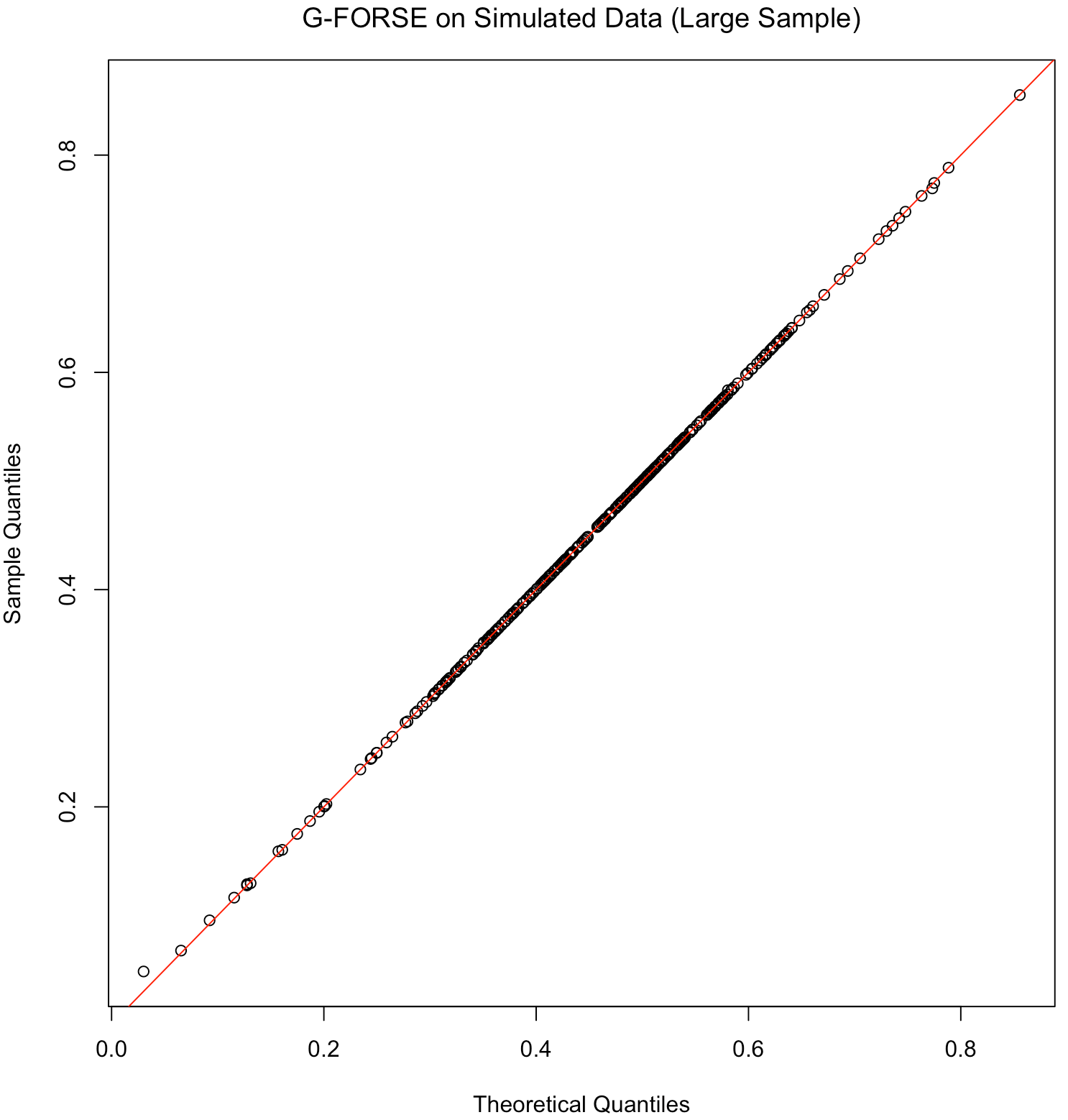}
\caption{}
\end{subfigure}
\begin{subfigure}[b]{0.24\textwidth}
\includegraphics[width=\textwidth]{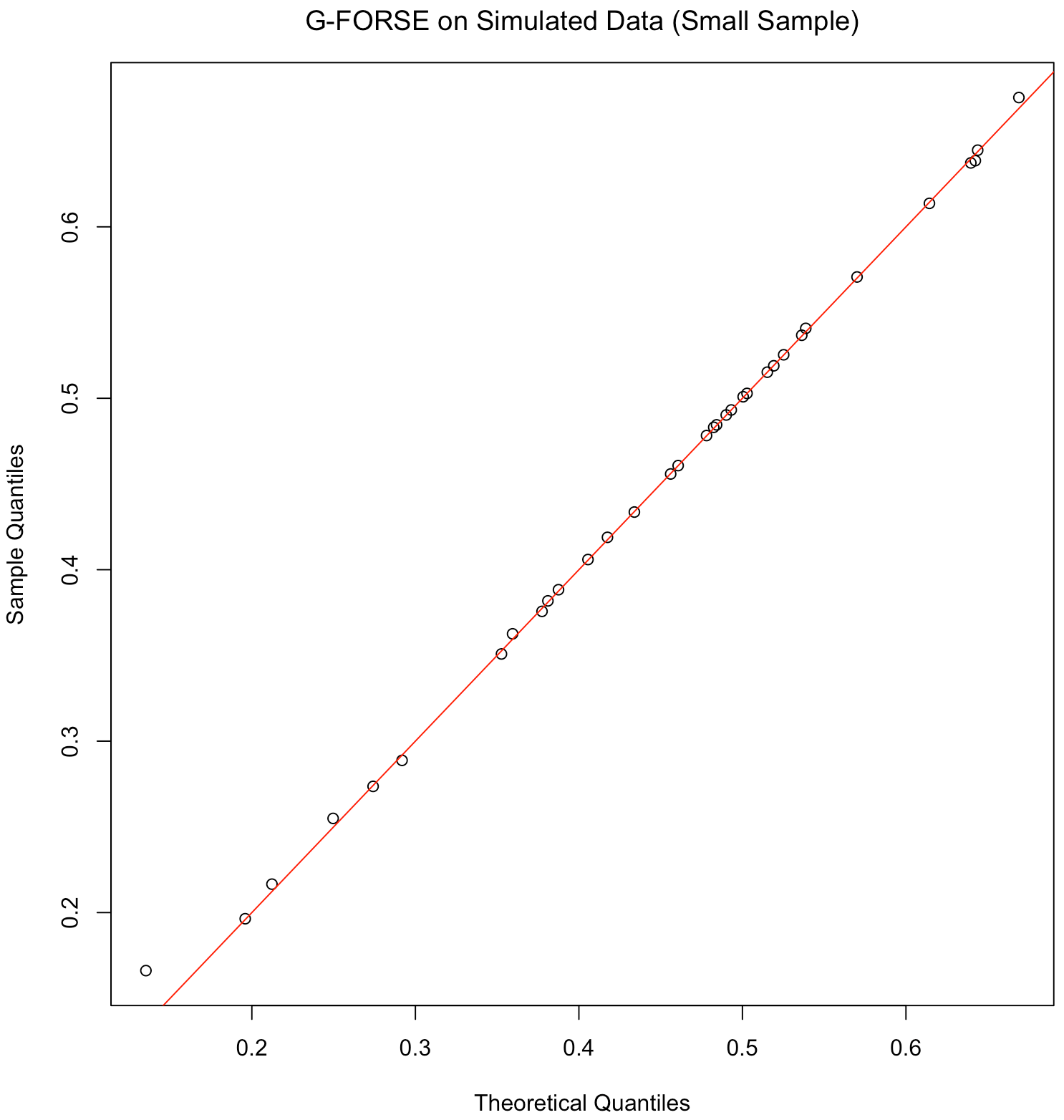}
\caption{}
\end{subfigure}
\begin{subfigure}[b]{0.24\textwidth}
\includegraphics[width=\textwidth]{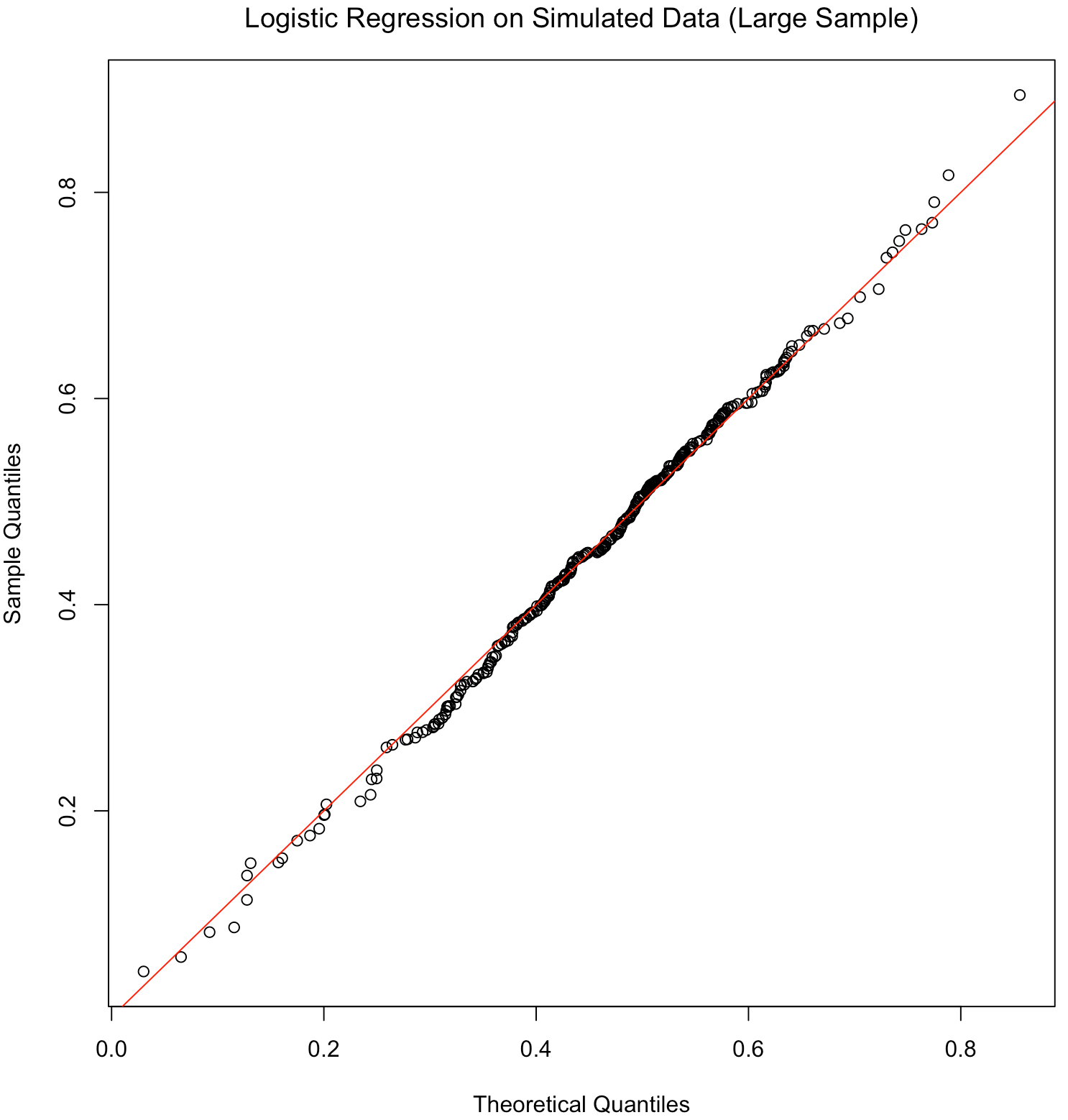}
\caption{}
\end{subfigure}
\begin{subfigure}[b]{0.24\textwidth}
\includegraphics[width=\textwidth]{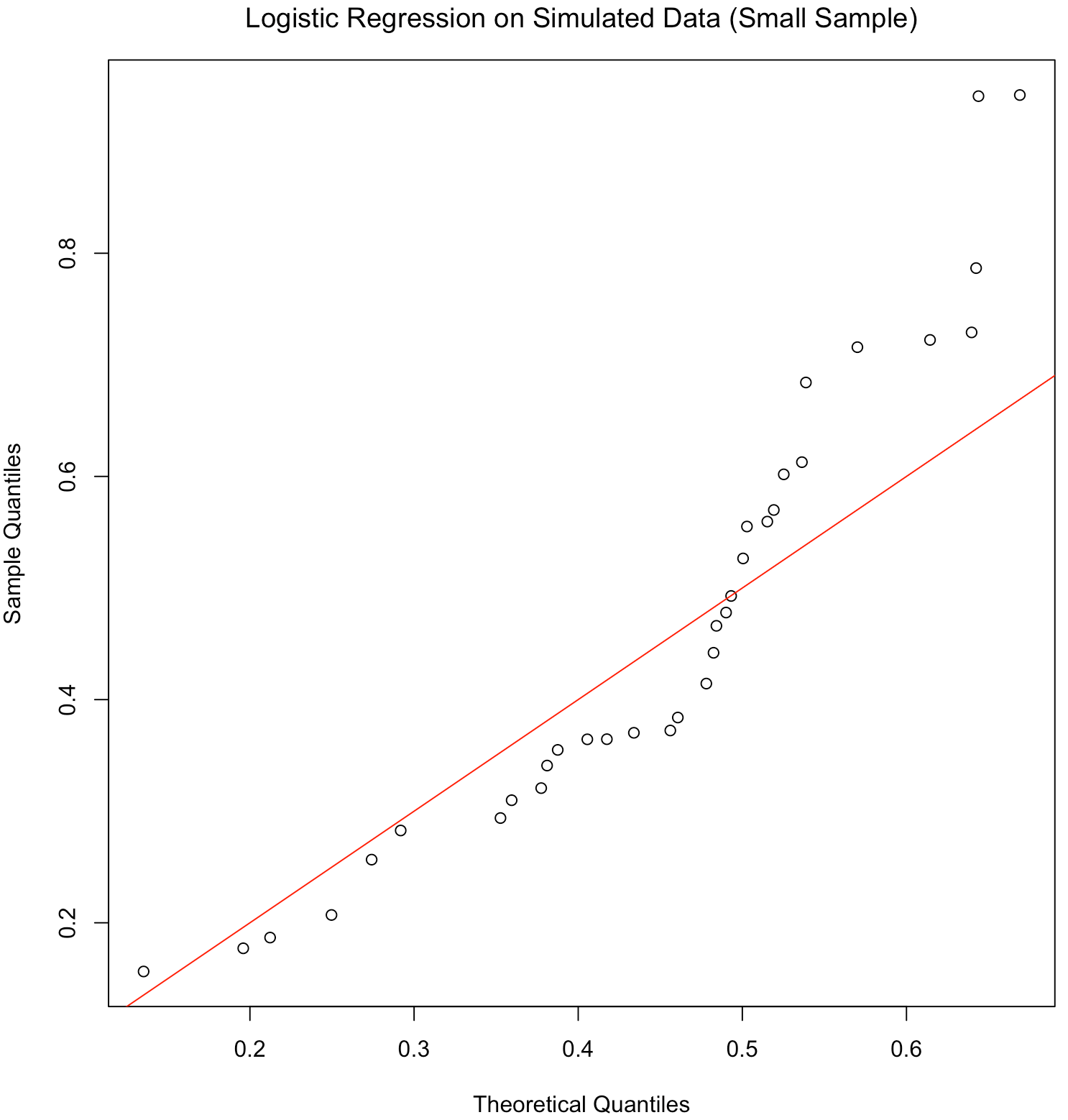}
\caption{}
\end{subfigure}
\caption{Scatter plots of predicted outcomes computed by G-forse and Logistic Regression on the simulated data versus the true values. G-forse appears to be superior to Logistic Regression in small samples.}\label{fig:qqGLR}
    \end{figure*}

\subsection{G-forse Validation}
We implement two synthetic experiments with the goal of evaluating the accuracy of model interpretation provided by the G-forse algorithm.
In both experiments, we apply G-forse (Section \RN{3}) on a ground-truth multi-variate function $f(x)$ to fit a metamodel $g(x)\approx f(x)$, and compare the resulting estimated coefficients for $g(x)$ against the true generator function $f(x)$.

In Figure \ref{fig:barplots}(a), we compare G-forse (blue) and Logistic Regression (yellow) in terms of the coefficients they estimate and their goodness-of-fit with different sample sizes with respect to the true function (Figure \ref{fig:qqGLR}).
We consider a true GLM function with five linear terms $X_1,\ldots,X_5$ and one interaction term $X_2X_3$.
As we can see, G-forse is generally more accurate than Logistic Regression. 
Moreover, G-forse tends to be more robust than Logistic Regression in the sense that Hauck-Donner effect would not degrade G-forse estimation, but it does so on Logistic Regression.
The Hauck-Donner effect (HDE) occurs when a Wald test statistic is not uniformly increasing as a function of separation between the estimated parameter and the null input. 
This is vividly visible in Figure \ref{fig:barplots}(a) where the yellow bars representing Logistic Regression failed to capture true coefficients (green bars) correctly.

%% file: tex/6-advantage.tex
This new machine learning interpretation technique comes with advantages and disadvantages relative to previous interpretation methods.
The disadvantages are primarily that G-forse may not be scalable on ultra-high dimensional data, and data smoothness is necessary to obtain a reliable estimation. 
Similar to other techniques, multicollinearity can affect explanation negatively.
The advantages are that black-box models can be interpreted with fewer parameters, missing data can be interpolated, a wide variety of covariance functions can be plugged into the process, it is applicable on noisy observations, and it can provide uncertainty estimate such as expected improvement.  
One can also benefit from using G-forse on cases with no maximum likelihood estimations as it was discussed earlier for Logistic Regression.

The G-forse models may also gain some statistical advantage from the estimate of uncertainty for not being based on pre-assumed models, but rather depending on empirical observations, which makes it superior to linear models or similar interpolation techniques.
Another advantage is a tendency of G-forse to be less varied toward specific bias direction, and the estimations are best linear unbiased estimator (BLUE) if the observations are spatially independent.
Additionally, G-forse can replace interpretable GLMs such as Logistic Regression suffering from HDE.

\subsection{How G-forse Interprets Better?}
The actual problem within black-box interpretation is to determine the 2-D relationship between features and samples themselves. 
A few extra steps are required to reveal the underlying relationships between features and samples whereas none of the existing interpretable methods provides this relationship.
If we are better aware of the connection between the instances and their networks with outcome, we would have a better interpretation about a black-box that could predict a complex underlying system, as such a model interpreter would align with the actual explanation within the black-box model.

Furthermore, it would have benefited us to appreciate the importance of a metamodel being responsive to nonlinear relationship.
LIME (as a popular interpreter) assumes that any nonlinear model can be explained linearly at a local space.
Obviously, this assumption is inappropriate if the network of variables are ignored in the explanation task (e.g., a network of variables may have different connections across data points).

Less obvious, the existing interpretable methods sacrifice accuracy for less complexity to ensure that the black-box model would be explained clearly by the interpreter.
Because of the incorporation of metamodeling techniques in interpretation problems, we gain more access to powerful and accepted techniques. 
This results in models with higher accuracy of prediction that can be the same as (or at least comparable to) the underlying complex model, while preserving the interpretability.
Lastly, the Hauck-Donner Effect as a hidden phenomenon can afflict many types of regression interpretable models, which leads to unreliable explanations.
Knowing all of this compelled us to introduce metamodeling as a new gateway to black-box interpretation.



%% file: tex/7-conc.tex
A complicated machine learning model is an abstraction of discovered patterns in massive data; a metamodel is a simplified version, reflecting properties of that complex model.

The main goal of this work was to propose a new technique based on metamodeling concept for turning a black-box into a clear-box. 
Metamodeling is widely used in simulation and engineering fields for covering one aspect of a modeling problem, and this study showed that it can be applied for machine learning interpretation as well.


Many straightforward extensions are suggested by this study:
in the explanations for image classifiers, G-forse framework can be extended to \textit{highlight the super-pixels} with the relatedness measurement afforded by variograms (calibrated by lag, sill, range and nugget) towards a specific class, to pictorialize why a certain prediction would happen.
The covariance function used in G-forse can be replaced by other types of functions, and G-forse can be scalable for ultra-high dimensional data using penalized likelihood function mixed with feature selection techniques \cite{joseph2008blind}.

This article has presented the viability of metamodeling in interpreting black-box models, suggesting that these research directions could prove useful.